# DenSe-AdViT: A novel Vision Transformer for Dense SAR Object Detection

Yang Zhang
*School of Artificial Intelligence*
*Beijing University of Posts and Telecommunications*
Beijng, China
zhangyang0706@bupt.edu.cn

Jingyi Cao[*]
*School of Artificial Intelligence*
*Beijing University of Posts and Telecommunications*
Beijng, China
caojingyi@bupt.edu.cn

Yanan You[*]
*School of Artificial Intelligence*
*Beijing University of Posts and Telecommunications*
Beijng, China
youyanan@bupt.edu.cn

Yuanyuan Qiao
*School of Artificial Intelligence*
*Beijing University of Posts and Telecommunications*
Beijng, China
yyqiao@bupt.edu.cn

*Abstract*—Vision Transformer (ViT) has achieved remarkable results in object detection for synthetic aperture radar (SAR) images, owing to its exceptional ability to extract global features. However, it struggles with the extraction of multi-scale local features, leading to limited performance in detecting small targets, especially when they are densely arranged. Therefore, we propose Density-Sensitive Vision Transformer with Adaptive Tokens (DenSe-AdViT) for dense SAR target detection. We design a Density-Aware Module (DAM) as a preliminary component that generates a density tensor based on target distribution. It is guided by a meticulously crafted objective metric, enabling precise and effective capture of the spatial distribution and density of objects. To integrate the multi-scale information enhanced by convolutional neural networks (CNNs) with the global features derived from the Transformer, Density-Enhanced Fusion Module (DEFM) is proposed. It effectively refines attention toward target-survival regions with the assist of density mask and the multiple sources features. Notably, our DenSe-AdViT achieves 79.8% mAP on the RSDD dataset and 92.5% on the SIVED dataset, both of which feature a large number of densely distributed vehicle targets.

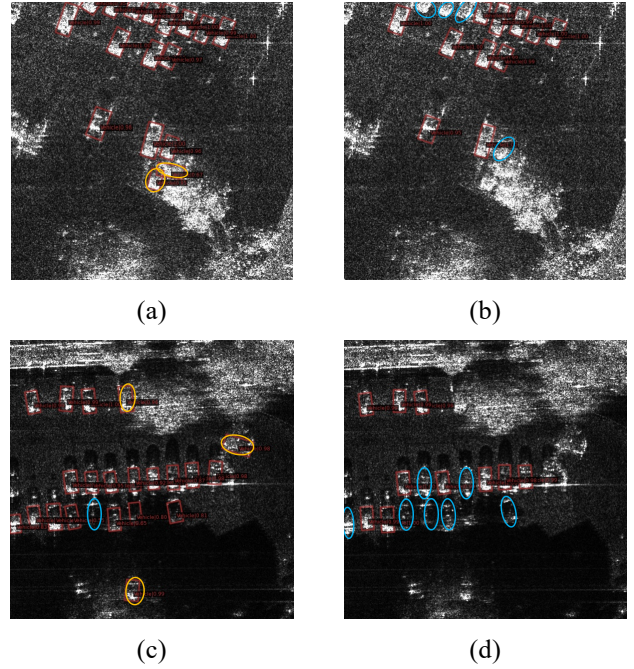

Fig. 1. Detection results on SAR images with dense target distributions using Oriented-RCNN [7] and ViT. (a) and (c) correspond to CNN-based approaches, while (b) and (d) represent transformer-based approaches. Yellow oval boxes indicate false alarms, blue ones indicate missed detections.

## I. INTRODUCTION

Detecting targets in synthetic aperture radar (SAR) images presents challenges due to the complex and cluttered nature of the scenes. In particular, factors such as speckle noise and signal scattering make it difficult to distinguish individual targets in SAR images with densely arranged objects. From the perspectives of high accuracy, low latency [1], and lightweight design [2], Convolutional Neural Networks (CNNs) have achieved significant success in SAR target detection with its strong capability in local feature extraction. However, struggling with long-range dependencies and global context, great potential for accuracy untapped is left in the application of dense SAR image target detection. With the rapid advancement of the Detection Transformer [3] and Vision Transformer (ViT), the attention-enhanced variants, such as OEGR-DETR [4] LRTransDet [5] and contrastive learning-based ViTs [6], have demonstrated remarkable adaptability to SAR image sensitive to imaging direction and significantly affected by scattering interference. These advancements further underscore the potential of Transformer architectures, which incorporate global environmental or scattering condition information, in addressing increasingly complex SAR target detection tasks.

However, as depicted in Fig. 1, both ViTs and CNNs exhibit limitations when dealing with small, dense, and overlapping targets in SAR images. Specifically, CNNs tend to produce false positives in non-target areas (such as grass or mountains), achieving higher recall rates for multi-scale dense targets. In contrast, ViTs demonstrate greater accuracy in dense regions but miss certain targets. Logically, it can be attributed to the interplay of the convolution under multilevel receptive field and multi-head attention mechanisms.

As a matter of fact, the catalyst for high-precision detection of small-scale SAR targets lies in identifying and focusing on dense target locations and then making purposeful feature extraction. Building on this concept, we propose Density-Sensitive Vision Transformer with Adaptive Tokens (DenSe-AdViT). Inspired by hybrid network architectures, as the supportive process features, the outputs of the CNN serve, on one hand, as the attention controller for salient regions, guiding the subsequent attention block to adaptively select and fine-tune information across tokens. On the other hand, the CNN introduces multi-scale information to compensate for the coarse feature that arises from the single-scale ViT extractor. The selection of token embeddings can be conceptualized as a focused attention mechanism on target region. It enhances the extraction of small-scale target information while effectively suppressing false positives. In the prediction phase, it ensures both efficient and accurate detection of dense, small targets while significantly reducing computational cost.

This work was supported by the National Key R&D Program of China (2023YFC3305901) and the BUPT innovation and entrepreneurship support program (2025-YC-A059, 2025-YC-S005). Corresponding authors: Jingyi Cao, Email: caojingyi@bupt.edu.cn.; Yanan You, Email: youyanan@bupt.edu.cn.
[*]corresponding author

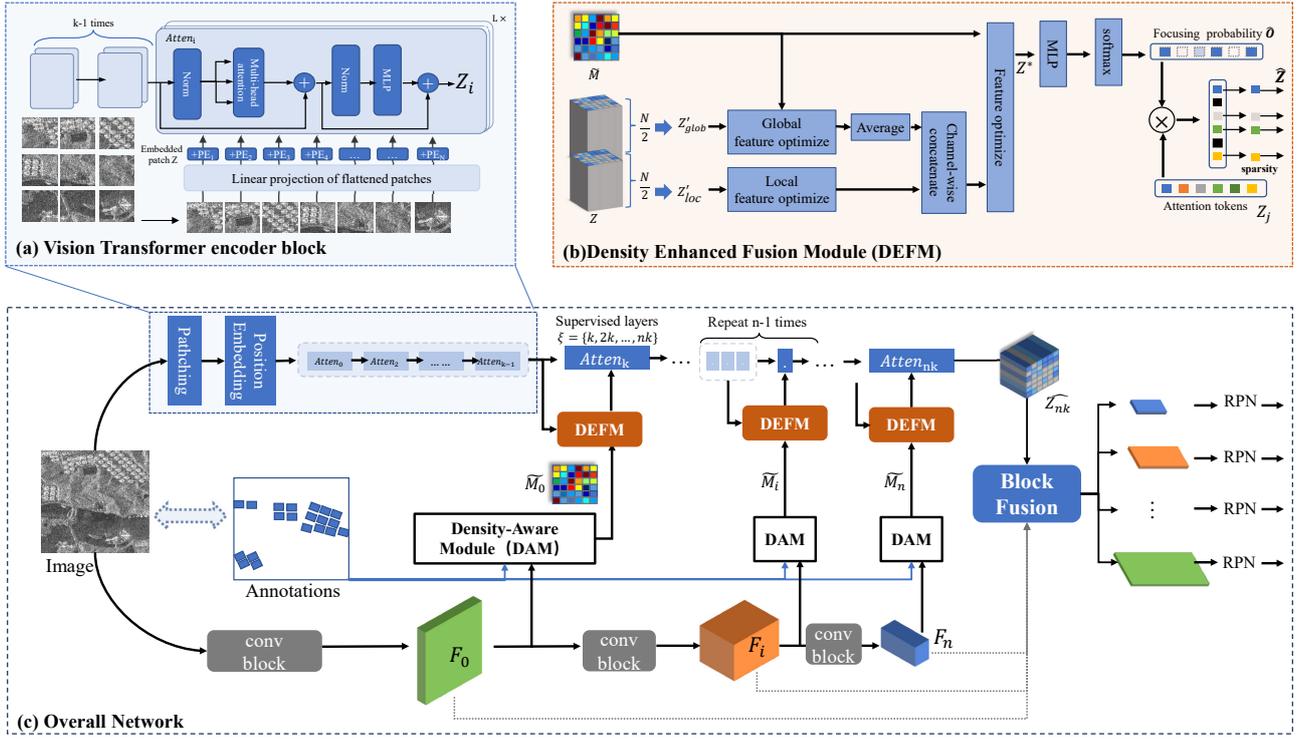

**Fig. 2.** Overview network of the proposed Density-Sensitive Vision Transformer with Adaptive Tokens.

## II. METHOD

### A. Overview

Building upon the classical ViT architecture, DenSe-AdViT incorporates continuous attention modules as the backbone while interleaving Density Enhanced Fusion Modules (DEFM) before the specified attention layers $l \in \xi$. The DEFM leverages sparse embedding mechanisms to focus on dense target regions. Specifically, we first partition the input image into patches, transform them into tokens, and add positional embeddings to update the tokenized features. The combined features are then fed into the attention branch. To explicitly identify dense target regions, we introduce a Density Aware Module (DAM), which generates a coarse density mask $\mathcal{M}$ with the annotation of targets. Simultaneously, a lightweight CNN architecture is employed to extract multi-scale convolutional features $F$, guiding the refinement of the initial density mask into an updated version $\widetilde{\mathcal{M}_i}$. Subsequently, DEFM takes the scene mask and the current stage's token embeddings as inputs. After several convolution operations, a token focus tensor is produced, which selectively filters the current stage's token embeddings, yielding a sparsified attentional feature $attn_i$. Upon completing the attention refinement across multiple DEFM-assisted stages, the final multi-level features from the attention mechanism are integrated with the multi-scale CNN outputs, forming the input for the multi-level Region Proposal Network (RPN). Through Intersection over Union (IoU)-based selection and position regression, the detection accuracy for small-scale SAR targets under complex backgrounds is significantly improved. These modules are described in more details below.

### B. Vision Transformer Backbone

Given an input SAR image $X \in \mathbb{R}^{H \times W}$, where $H$ and $W$ denote the height and width of the image respectively. Firstly, we divide the image into $N$ blocks with size $P \times P$, and map them into D-dimensional vector $z_i \in \mathbb{R}^D$ with linear projection. The vectors of all the image blocks are stacked to obtain a matrix of $N \times D$, representing the representation of all the blocks. Next, position information is added with the token embedding by position encoding $PE_i \in \mathbb{R}^D$ to output the block representation $Z \in \mathbb{R}^{N \times D}$.

$$Z = [z_1 + PE_1, z_2 + PE_2, \ldots, z_N + PE_N] \quad (1)$$

The block representation $Z$ is passed as input to the multilayer Transformer encoder. Each layer of the encoder consists of a multi-head self-attention mechanism and a feed-forward neural network. Given supervised layers $\xi$, for every $i$ ($i \notin \xi$) layer encoder:

$$Q_i = Z_i \cdot W_Q^i, K_i = Z_i \cdot W_K^i, V_i = Z_i \cdot W_V^i \quad (2)$$

$$Z_i' = \text{MultiHeadAttention}(Q^i, K^i, V^i) + Z_i \quad (3)$$

$$Z^{i+1} = LN\left(\text{FFN}\left(LN\left(Z_i'\right)\right) + Z_i'\right) \quad (4)$$

where $Q_i, K_i, V_i$ are the Query, Key and Value obtained from the input representation by linear transformation, respectively. FFN represents feed forward network and LN represents layer normalization. It is particularly noteworthy that, for the layer $j$ ($j \in \xi$), the DEFM was performed. The output mask $\widehat{O}$ of DEFM is produced to highlight the salient tokens and update the embedding of each token. The update process of $Z_j$ is formulated as the follows:

$$Z_j \leftarrow Z_j \odot \widehat{O} \quad (5)$$

Additionally, in the final stage, the updated $Z_n$ undergoes token position rearrangement and is fused with $F_n$, resulting in multi-level features to participate the subsequent region proposal and ultimately produce the detection results.

## C. Density Aware Module

This module generates density-aware masks that reflect the concentration and spatial arrangement of target objects within an image. These masks are constructed by leveraging ground truth bounding boxes and employing Gaussian kernels to model the density distribution around each object's center.

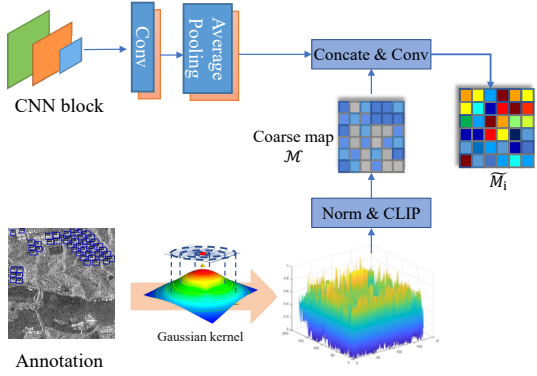

**Fig. 3.** The details of density aware module

After initializing the scene density map with $\mathcal{M} \in \mathbf{0}^{H \times W}$. The ground truth bounding boxes are used to compute Gaussian kernels of each object. Define Gaussian kernel at a pixel $(x, y)$ as follows:

$$\mathcal{G}(x,y) = \exp\left(-\frac{(x-c_x)^2 + (y-c_y)^2}{2\sigma^2}\right) \quad (6)$$

where $c_x$ and $c_y$ denote the center coordinates of the bounding box, $\sigma$ is the standard deviation calculated with height, weight and angle of the bounding box to control the spatial extent of the density influence. By aggregating the Gaussian contributions with all $N$ annotated objects, the scene density map is then computed as:

$$\mathcal{M}(x,y) = \sum_{i=1}^{N} \mathcal{G}_i(x,y) \quad (7)$$

This summation ensures that regions with a higher concentration of objects exhibit stronger density signals, while areas with fewer objects have weaker signals.

But it only provides a coarse density-aware representation. Therefore, for each block awaiting density fusion, we perform feature integration and secondary refinement based on the outputs $F_i$ from the four-level convolutional neural network. This process generates a density mask with multi-level semantic adaptability, while effectively mitigating the crude operations of the coarse mask that can lead to feature loss in sparse target regions.

$$\widetilde{\mathcal{M}}_\iota = \begin{cases} conv\big(concat(\mathcal{M}, AveragePool(F_i))\big), trainning \\ conv\big(AveragePool(F_i)\big), inferring \end{cases} \quad (8)$$

To normalize and ensure mask's values lie within the range [0,1], a clipping operation is applied to output robust spatial prior to characterize object distribution.

$$\widetilde{\mathcal{M}}_\iota(x,y) = \text{clip}\big(\widetilde{\mathcal{M}}_\iota(x,y), 0, 1\big) \quad (9)$$

## D. Density Enhanced Fusion Module

DEFM aims at enhancing the focus on regions of interest by effectively integrating the local structural information from CNN, the globally aggregated features from ViT, and the density-aware mask $\widetilde{\mathcal{M}}_\iota$. Formally, let the input feature tensor be denoted as $Z_j \in \mathbb{R}^{B \times N \times C}$, where $B$ represents the batch size, $N$ represents the sequence length (corresponding to feature patches), and $C$ the number of channels.

Upon receiving $Z_j$, DEFM first processes it through a series of normalization and non-linear transformation layers, e.g. LN and GELU. To balance the extraction of local and global representations, $Z_j$ is partitioned along the channel dimension into two distinct components:

$$Z'_{loc} \in \mathbb{R}^{B \times N \times \frac{C}{2}} \quad Z'_{glob} \in \mathbb{R}^{B \times N \times \frac{C}{2}}, \quad (10)$$

where $Z'_{loc}$ retains localized contextual information, and $Z'_{glob}$ encapsulates globally aggregated features.

To accentuate target regions and mitigate the influence of irrelevant background areas, DEFM reshape the density mask $\widetilde{\mathcal{M}}_\iota$ and broadcasts $\frac{C}{2}$ times in channel dimension to produce $P \in \mathbb{R}^{B \times N \times \frac{C}{2}}$. It modulates global features with the element-wise multiplication followed by the weighted summation across the token sequence dimension:

$$\mathcal{g} = \frac{\sum_{n=1}^{N} Z'^{(n)}_{glob} \odot P^{(n)}}{\sum_{n=1}^{N} P^{(n)}} \quad (11)$$

where $Z'^{(n)}_{glob}$ and $P^{(n)}$ denote the feature vectors at the $n$-th token position, $\odot$ represents element-wise multiplication. The denominator ensures normalization, preventing the global vector $\mathcal{g}$ from being disproportionately scaled due to varying sequence lengths or target densities.

The resultant global vector $\mathcal{g}$ is then broadcasted to concatenate with $Z'_{loc}$ along the channel dimension to form the fused representation $Z'$:

$$Z' = \left[Z'_{loc}, \underbrace{\mathcal{g}, \mathcal{g}, \dots, \mathcal{g}}_{N \text{ times}}\right] \in \mathbb{R}^{B \times N \times C} \quad (12)$$

During the training phase, this fused feature $Z'$ is further modulated by the corresponding density mask $\widetilde{\mathcal{M}}_\iota$, suppressing the region respond that not associated with target distributions:

$$Z^* = Z' \odot \widetilde{\mathcal{M}}_\iota \quad (13)$$

where $Z^*$ represents the final feature tensor after mask application.

In order to make network pay more attention to the interest region, the processed features $Z^*_i$ are passed through the output convolutional branch, which comprises additional normalization, multiple linear transformations interleaved with non-linear activations, and a log-softmax layer to produce the final focusing probability $\widehat{\boldsymbol{O}} \in \mathbb{R}^{B \times N \times 2}$.

$$\widehat{\boldsymbol{O}} = softmax\big(MLP(Z^*_i)\big) \quad (14)$$

The focusing probability then acts on the tokens of attention to clarify the effectiveness in the sequence. Due to its implied target density, it can fully enable focused information extraction in dense regions.

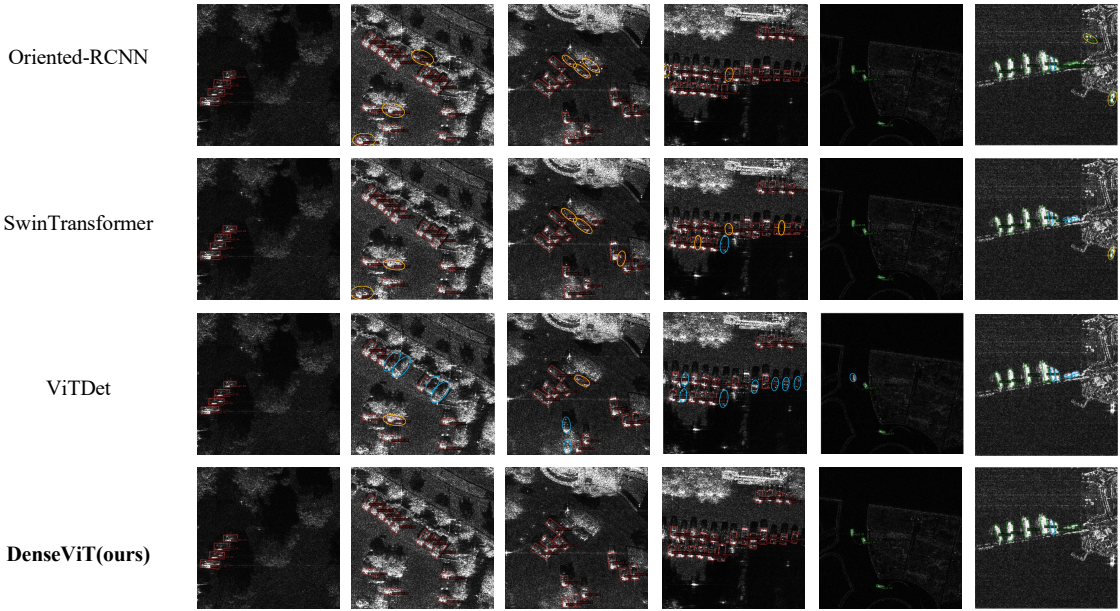

**Fig. 4.** Visualization of detection results on different methods. The first through fourth columns are densely packed vehicles (red boxes), the fifth to sixth columns show ships (green boxes). Yellow oval boxes indicate false alarms, blue oval boxes indicate missed detections.

## III. EXPERIMENT

### A. Datasets

We have conducted sufficient experiments on the RSDD and SIVED datasets. Rotated Ship Detection Dataset in SAR images (RSDD) [8] includes 7,000 slices and 10,263 ship instances of multi-observing modes, multi-polarization modes, and multi-resolutions. SAR image dataset for vehicle detection (SIVED) [9] uses Ka, Ku, and X bands of data and contains lots of dense targets. Both of the two datasets are well fitted for evaluating our algorithms. We used rotated box labeling to better capture the features of dense targets.

### B. Details

In detail, during data preprocessing, we resize all input images to 512×512 and apply multiple data augmentation strategies, including random flipping and cropping. In the attention branch, we employ 16×16 patches, multi-head attention with 8 heads, and a total depth of 12.

To ensure stable and effective training, we use the AdamW optimizer with an initial learning rate of $10^{-4}$, $\beta$ set to (0.9, 0.999), and a weight decay of 0.01, while excluding positional embeddings and normalization layers from decay. We also adopt gradient clipping to prevent exploding gradients. Furthermore, a cosine annealing schedule with a minimum learning rate of $10^{-6}$ is employed in conjunction with a linear warmup phase, where warmup-iters is 1000, ensuring a gradual learning rate ramp-up and enhanced training stability during the initial stages.

### C. Results

Using mean Average Precision (mAP) as the validation metric, our algorithm achieved SOTA results on both datasets, only slightly underperforming the CNN method in terms of recall rate. Fig. 4 shows visualization of detection results on the four different methods.

From the experimental results in Fig. 4 and Table. 1, it can be observed that the CNN-based approach, benefiting from its strong local feature perception, achieves a high recall rate for dense targets but tends to generate more false alarms, leading to comparatively lower accuracy. Among Transformer-based methods, SwinTransformer, despite leveraging a windowed attention mechanism, continues to face the inherent challenges of Transformers in handling dense target features. Conversely, ViTDet [11], owing to its enhanced feature extraction and global receptive field, attains higher accuracy but experiences a noticeable drop in recall.

**Table. 1.** Result metrics of the proposed method and others.

|  | RSDD | | SIVED | |
| --- | --- | --- | --- | --- |
|  | mAP | Recall | mAP | Recall |
| Oriented-RCNN [7] | 0.770 | 0.811 | 0.902 | **0.981** |
| SwinTransformer [11] | 0.784 | 0.839 | 0.901 | 0.976 |
| ViTDet [12] | 0.773 | 0.821 | 0.913 | 0.971 |
| **DenseViT(ours)** | **0.798** | **0.848** | **0.925** | 0.978 |

In contrast, our proposed method achieves an accuracy of 92.5% on the densely populated SIVED dataset—surpassing other approaches—and only incurs a minor reduction in recall compared to the CNN-based method. From the results, it can be seen that our method reduces the leakage of dense regions and false alarms in other regions compared to other methods due to the ability to observe dense regions effectively. Moreover, on the RSDD dataset, which contains both dense and sparse targets, our method demonstrates leading performance, indicating that it preserves robust detection capability under conventional conditions.

## IV. CONCLUSION

With proposing a Density-Aware Module (DAM) and a Density-Enhanced Fusion Module (DEFM), our method combines multi-scale CNN features with global attention，effectively dealt with the difficult problem of detecting dense targets. DenSe-AdViT demonstrates superior accuracy and recall, particularly in dense target scenarios, while maintaining robust performance in conventional conditions.

## REFERENCES

[1] C. Yu and Y. Shin, "An efficient YOLO for ship detection in SAR images via channel shuffled reparameterized convolution blocks and dynamic head," *ICT Express*, 2024.

[2] X. Xu, X. Zhang, and T. Zhang, "Lite-YOLOv5: A Lightweight Deep Learning Detector for On-Board Ship Detection in Large-Scene Sentinel-1 SAR Images," *Remote Sensing*, vol. 14, no. 4, 2022.

[3] N. Carion, F. Massa, G. Synnaeve, N. Usunier, A. Kirillov, and S. Zagoruyko, "End-to-End Object Detection with Transformers," *CoRR*, vol. abs/2005.12872, 2020.

[4] Y. Feng, Y. You, J. Tian, and G. Meng, "OEGR-DETR: A Novel Detection Transformer Based on Orientation Enhancement and Group Relations for SAR Object Detection," *Remote Sensing*, vol. 16, no. 1, p. 106, 2024.

[5] K. Feng, L. Lun, X. Wang, and X. Cui, "LRTransDet: A Real-Time SAR Ship-Detection Network with Lightweight ViT and Multi-Scale Feature Fusion," *Remote Sensing*, vol. 15, no. 22, 2023.

[6] J. Deng, Y. Zhu, S. Zhang and S. Chen, "SAR Image Recognition Using ViT Network and Contrastive Learning Framework With Unlabeled Samples," *IEEE Geoscience and Remote Sensing Letters*, vol. 21, pp. 1-5, 2024

[7] Xie X, Cheng G, Wang J, et al. "Oriented R-CNN for object detection" *Proceedings of the IEEE/CVF international conference on computer vision*. 2021: 3520-3529.

[8] XU Congan, SU Hang, LI Jianwei, et al. RSDD-SAR: Rotated Ship Detection Dataset in SAR Images. *Journal of Radars*, 2022, 11(4): 581-599. doi: 10.12000/JR22007

[9] X. Lin, B. Zhang, F. Wu, C. Wang, Y. Yang, and H. Chen, "SIVED: A SAR Image Dataset for Vehicle Detection Based on Rotatable Bounding Box," *Remote Sensing*, vol. 15, no. 11, 2023, doi: 10.3390/rs15112825.

[10] Xie X, Cheng G, Wang J, et al. "Oriented R-CNN for object detection" *Proceedings of the IEEE/CVF international conference on computer vision*. 2021: 3520-3529.

[11] Z. Liu et al., "Swin Transformer: Hierarchical Vision Transformer using Shifted Windows," *Proceedings of the IEEE/CVF international conference on computer vision.* 2021: 10012-10022.

[12] Y. Li, H. Mao, R. B. Girshick, and K. He, "Exploring Plain Vision Transformer Backbones for Object Detection," *European conference on computer vision.* 2022: 280-296.